\documentclass[11pt,a4paper]{article}
\usepackage[hyperref]{acl2020}
\usepackage{times}
\usepackage{latexsym}

\usepackage{booktabs}
\usepackage{graphicx}
\usepackage{wtref}
\usepackage{amsmath,bm}
\usepackage{amsfonts}
\usepackage{etoolbox}
\usepackage{url}

\usepackage{microtype}

\aclfinalcopy 
\def\aclpaperid{***} 

\usepackage{gb4e}
\noautomath

\makeatletter
\renewcommand{\@subex}[2]{\settowidth{\labelwidth}{#1}\itemindent\z@\labelsep#2%
         \topsep0\p@\itemsep0\p@%
         \parsep\p@\partopsep0\p@%
         \leftmargin\labelwidth%
         \ifnum\the\@xnumdepth=1
         \else\advance\leftmargin#2\relax\fi}
\makeatother

\newref{eq,fig,tab,sec}

\title{An Analysis of the Utility of Explicit Negative Examples \\ to Improve the Syntactic Abilities of Neural Language Models}

\author{Hiroshi Noji \\
  Artificial Intelligence Research Center \\
  AIST, Tokyo, Japan \\
  {\tt hiroshi.noji@aist.go.jp} \\\And
  Hiroya Takamura \\
  Artificial Intelligence Research Center \\
  AIST, Tokyo, Japan \\
  {\tt takamura.hiroya@aist.go.jp} \\}

\date{}

\begin{document}
\maketitle
\begin{abstract}
 We explore the utilities of explicit negative examples in training neural language models.
  Negative examples here are incorrect words in a sentence, such as {\it barks} in *{\it The dogs barks}.
 Neural language models are commonly trained only on positive examples, a set of sentences in the training data, but recent studies suggest that the models trained in this way are not capable of robustly handling complex syntactic constructions, such as long-distance agreement.
 In this paper, we first demonstrate that appropriately using negative examples about particular constructions (e.g., subject-verb agreement) will boost the model's robustness on them in English, with a negligible loss of perplexity.
 The key to our success is an additional margin loss between the log-likelihoods of a correct word and an incorrect word.
 We then provide a detailed analysis of the trained models.
 One of our findings is the difficulty of object-relative clauses for RNNs.
 We find that even with our direct learning signals the models still suffer from resolving agreement across an object-relative clause.
 Augmentation of training sentences involving the constructions somewhat helps, but the accuracy still does not reach the level of subject-relative clauses.
 Although not directly cognitively appealing, our method can be a tool to analyze the true architectural limitation of neural models on challenging linguistic constructions.
\end{abstract}

\section{Introduction}
\seclabel{intro}

Despite not being exposed to explicit syntactic supervision, neural language models (LMs), such as recurrent neural networks, are able to generate fluent and natural sentences, suggesting that they induce syntactic knowledge about the language to some extent.
However, it is still under debate whether such induced knowledge about grammar is robust enough to deal with syntactically challenging constructions such as long-distance subject-verb agreement.
So far, the results for RNN language models (RNN-LMs) trained only with raw text are overall negative;
prior work has reported low performance on the challenging test cases \cite{marvin-linzen:2018:EMNLP} even with the massive size of the data and model \cite{van-schijndel-EtAl:2019:EMNLP1},
or argue the necessity of an architectural change to track the syntactic structure explicitly \cite{wilcox-etal-2019-structural,kuncoro-EtAl:2018:Long}.
Here the task is to evaluate whether a model assigns a higher likelihood on a grammatically correct sentence (\ref{ex:correct_orc}) over an incorrect sentence (\ref{ex:incorrect_orc}) that is minimally different from the original one \cite{Q16-1037}.

\vspace{2pt}\begin{exe}
 \ex\label{ex:orc}
 \begin{xlist}
  \ex[]{\label{ex:correct_orc}The author that the guards like \underline{laughs}.}
  \ex[*]{\label{ex:incorrect_orc}{The author that the guards like \underline{laugh}.}}
 \end{xlist}
\end{exe}
\vspace{2pt}

In this paper, to obtain a new insight into the syntactic abilities of neural LMs, in particular RNN-LMs, we perform a series of experiments under a different condition from the prior work.
Specifically, we extensively analyze the performance of the models that are exposed to explicit negative examples.
In this work, negative examples are the sentences or tokens that are grammatically incorrect, such as (\ref{ex:incorrect_orc}) above.

Since these negative examples provide a direct learning signal on the task at test time it may not be very surprising if the task performance goes up.
We acknowledge this, and argue that our motivation for this setup is to deepen understanding, in particular the limitation or the capacity of the current architectures, which we expect can be reached with such strong supervision.
Another motivation is engineering:
we could exploit negative examples in different ways, and establishing a better way will be of practical importance toward building an LM or generator that can be robust on particular linguistic constructions.

The first research question we pursue is about this latter point:
what is a better method to utilize negative examples that help LMs to acquire robustness on the target syntactic constructions?
Regarding this point, we find that adding additional token-level loss trying to guarantee a margin between log-probabilities for the correct and incorrect words (e.g., $\log p(\textrm{laughs} | h)$ and $\log p(\textrm{laugh} | h)$ for (\ref{ex:correct_orc})) is superior to the alternatives.
On the test set of \citet{marvin-linzen:2018:EMNLP}, we show that LSTM language models (LSTM-LMs) trained by this loss reach near perfect level on most syntactic constructions for which we create negative examples, with only a slight increase of perplexity about 1.0 point.

Past work conceptually similar to us is \citet{enguehard-etal-2017-exploring}, which, while not directly exploiting negative examples, trains an LM with additional explicit supervision signals to the evaluation task.
They hypothesize that LSTMs do have enough capacity to acquire robust syntactic abilities but the learning signals given by the raw text are weak, and show that multi-task learning with a binary classification task to predict the upcoming verb form (singular or plural) helps models aware of the target syntax (subject-verb agreement).
Our experiments basically confirm and strengthen this argument, with even stronger learning signals from negative examples, and we argue this allows us to evaluate the true capacity of the current architectures.
In our experiments (Section~\secref{exp}), we show that our margin loss achieves higher syntactic performance than their multi-task learning.

Another relevant work on the capacity of LSTM-LMs is \citet{kuncoro-etal-2019-scalable}, which shows that by distilling from syntactic LMs \cite{dyer-EtAl:2016:N16-1}, LSTM-LMs can improve their robustness on various agreement phenomena.
We show that our LMs with the margin loss outperform theirs in most of the aspects, further strengthening the argument about a stronger capacity of LSTM-LMs.

The latter part of this paper is a detailed analysis of the trained models and introduced losses.
Our second question is about the true {\it limitation} of LSTM-LMs:
are there still any syntactic constructions that the models cannot handle robustly even with our direct learning signals?
This question can be seen as a fine-grained one raised by \citet{enguehard-etal-2017-exploring} with a stronger tool and improved evaluation metric.
Among tested constructions, we find that syntactic agreement across an object relative clause (RC) is challenging.
To inspect whether this is due to the architectural limitation, we train another LM on a dataset, on which we unnaturally augment sentences involving object RCs.
Since it is known that object RCs are relatively rare compared to subject RCs \cite{hale-2001-probabilistic},
frequency may be the main reason for the lower performance.
Interestingly, even when increasing the number of sentences with an object RC by eight times (more than twice of sentences with a subject RC), the accuracy does not reach the same level as agreement across a subject RC.
This result suggests an inherent difficulty in tracking a syntactic state across an object RC for sequential neural architectures.

We finally provide an ablation study to understand the encoded linguistic knowledge in the models learned with the help of our method.
We experiment under reduced supervision at two different levels:
(1) at a lexical level, by not giving negative examples on verbs that appear in the test set;
(2) at a construction level, by not giving negative examples about a particular construction, e.g., verbs after a subject RC.
We observe no huge score drops by both.
This suggests that our learning signals at a lexical level (negative words) strengthen the abstract syntactic knowledge about the target constructions, and also that the models can generalize the knowledge acquired by negative examples to similar constructions for which negative examples are not explicitly given.
The result also implies that negative examples do not have to be complete and can be noisy, which will be appealing from an engineering perspective.

\section{Target Task and Setup}
The most common evaluation metric of an LM is perplexity.
Although neural LMs achieve impressive perplexity \cite{merity2018regularizing}, it is an average score across all tokens and does not inform the models' behaviors on linguistically challenging structures, which are rare in the corpus.
This is the primary motivation to separately evaluate the models' syntactic robustness by a different task.

\subsection{Syntactic evaluation task}
\seclabel{task}
As introduced in Section~\secref{intro}, the task for a model is to assign a higher probability to the grammatical sentence over the ungrammatical one, given a pair of minimally different sentences at a critical position affecting the grammaticality.
For example, (\ref{ex:correct_orc}) and (\ref{ex:incorrect_orc}) only differ at a final verb form, and to assign a higher probability to (\ref{ex:correct_orc}), models need to be aware of the agreement dependency between {\it author} and {\it laughs} over an RC.

\paragraph{\citet{marvin-linzen:2018:EMNLP} test set}
While initial work \cite{Q16-1037,N18-1108} has collected test examples from naturally occurring sentences, this approach suffers from the coverage issue, as syntactically challenging examples are relatively rare.
We use the test set compiled by \citet{marvin-linzen:2018:EMNLP}, which consists of synthetic examples (in English) created by a fixed vocabulary and a grammar.
This approach allows us to collect varieties of sentences with complex structures.

The test set is divided by the syntactic constructions appearing in each example.
Many constructions are different types of subject-verb agreement, including local agreement on different sentential positions (\ref{ex:simple}), and non-local agreement across different types of phrases.
Intervening phrases include prepositional phrases, subject RCs, object RCs, and coordinated verb phrases (\ref{ex:long}).
(\ref{ex:orc}) is an example of agreement across an object RC.
\vspace{2pt}\begin{exe}
 \ex\label{ex:simple}The senators \underline{smile/*smiles}.
\end{exe}
\begin{exe}
 \ex\label{ex:long}The senators like to watch television shows and \underline{are/*is} twenty three years old.
\end{exe}
\vspace{2pt}
Previous work has shown that non-local agreement is particularly challenging for sequential neural models \cite{marvin-linzen:2018:EMNLP}.

The other patterns are reflexive anaphora dependencies between a noun and a reflexive pronoun (\ref{ex:reflexive}), and
on negative polarity items (NPIs), such as {\it ever}, which requires a preceding negation word (e.g., {\it no} and {\it none}) at an appropriate scope (\ref{ex:npi}):
\vspace{2pt}\begin{exe}
 \ex\label{ex:reflexive}The authors hurt \underline{themselves/*himself}.
\end{exe}
\begin{exe}
 \ex\label{ex:npi}\underline{No/*Most} authors have {\it ever} been popular.
\end{exe}
\vspace{2pt}

Note that NPI examples differ from the others in that the context determining the grammaticality of the target word (No/*Most) does not precede it.
Rather, the grammaticality is determined by the following context.
As we discuss in Section~\secref{method}, this property makes it difficult to apply training with negative examples for NPIs for most of the methods studied in this work.

All examples above (\ref{ex:orc}--\ref{ex:npi}) are actual test sentences, and we can see that since they are synthetic some may sound somewhat unnatural.
The main argument behind using this dataset is that even not very natural, they are still strictly grammatical, and an LM equipped with robust syntactic abilities should be able to handle them as a human would do.

We use the original test set used in \citet{marvin-linzen:2018:EMNLP}.\footnote{We use the ``EMNLP2018'' templates in {\color{darkblue}https://github.com/BeckyMarvin/LM\_syneval}.}
See the supplementary materials of this for the lexical items and example sentences in each construction.

\subsection{Language models}
\seclabel{lm}
\paragraph{Training data}

Following the practice, we train LMs on the dataset not directly relevant to the test set.
Throughout the paper, we use an English Wikipedia corpus assembled by \citet{N18-1108}, which has been used as training data for the present task \cite{marvin-linzen:2018:EMNLP,kuncoro-etal-2019-scalable}, consisting of 80M/10M/10M tokens for training/dev/test sets.
It is tokenized and rare words are replaced by a single unknown token, amounting to the vocabulary size of 50,000.

\paragraph{Baseline LSTM-LM}
Since our focus in this paper is an additional loss exploiting negative examples (Section~\secref{method}), we fix the baseline LM throughout the experiments.
Our baseline is a three-layer LSTM-LM with 1,150 hidden units at internal layers trained with the standard cross-entropy loss.
Word embeddings are 400-dimensional, and input and output embeddings are tied \cite{DBLP:journals/corr/InanKS16}.
Deviating from some prior work \cite{marvin-linzen:2018:EMNLP,van-schijndel-EtAl:2019:EMNLP1},
we train LMs at sentence level as in sequence-to-sequence models \cite{sutskever2014sequence}.
This setting has been employed in some previous work \cite{kuncoro-EtAl:2018:Long,kuncoro-etal-2019-scalable}.\footnote{
On the other hand, the LSTM-LM of \citet{marvin-linzen:2018:EMNLP}, which is prepared by \citet{N18-1108}, is trained at document level through truncated backpropagation through time (BPTT) \cite{conf/icassp/MikolovKBCK11}.
Since our training regime is more akin to the task setting of syntactic evaluation, it may provide some advantage at test time.
}

Parameters are optimized by SGD.
For regularization, we apply dropout on word embeddings and outputs of every layer of LSTMs, with weight decay of 1.2e-6, and anneal the learning rate by 0.5 if the validation perplexity does not improve successively, checking every 5,000 mini-batches.
Mini-batch size, dropout weight, and initial learning rate are tuned by perplexity on the dev set of Wikipedia dataset.\footnote{Following values are found: mini-batch size: 128; initial learnin rate: 20.0; dropout weight on the word embedding layer and each output layer of LSTM: 0.1.}
Note that we tune these values for the baseline LSTM-LM and fix them across the experiments.

The size of our three-layer LM is the same as the state-of-the-art LSTM-LM at document-level \cite{merity2018regularizing}.
\citet{marvin-linzen:2018:EMNLP}'s LSTM-LM is two-layer with 650 hidden units and word embeddings.
Comparing two, since the word embeddings of our models are smaller (400 vs.~650) the total model sizes are comparable (40M for ours vs.~39M for theirs).
Nonetheless, we will see in the first experiment that our carefully tuned three-layer model achieves much higher syntactic performance than their model (Section~\secref{exp}), being a stronger baseline to our extensions, which we introduce next.

\section{Learning with Negative Examples}
\seclabel{method}

Now we describe four additional losses for exploiting negative examples.
The first two are existing ones, proposed for a similar purpose or under a different motivation.
As far as we know, the latter two have not appeared in past work.\footnote{
The loss for large-margin language models \cite{huang-EtAl:2018:EMNLP3} is similar to our sentence-level margin loss.
Whereas their formulation is more akin to the standard large-margin setting, aiming to learn a reranking model, our margin loss is simpler, just comparing two log-likelihoods of predefined positive and negative sentences.
}

We note that we create negative examples by modifying the original Wikipedia training sentences, not sentences in the test set.
As a running example, let us consider the case where sentence (\ref{ex:correct_batch}) exists in a mini-batch, from which we create a negative example (\ref{ex:incorrect_batch}).
\vspace{2pt}\begin{exe}
 \ex\label{ex:batch}
 \begin{xlist}
  \ex[]{\label{ex:correct_batch}An industrial park with several companies \underline{is} located in the close vicinity.}
  \ex[*]{\label{ex:incorrect_batch}An industrial park with several companies \underline{are} located in the close vicinity.}
 \end{xlist}
\end{exe}
\paragraph{Notations}
By a {\it target} word, we mean a word for which we create a negative example (e.g., {\it is}).
We distinguish two types of negative examples:
a {\it negative token} and a {\it negative sentence}; the former means a single incorrect word (e.g., {\it are}), while the latter means an entire ungrammatical sentence.

\subsection{Negative Example Losses}

\paragraph{Binary-classification loss}
This is proposed by \citet{enguehard-etal-2017-exploring} to complement a weak inductive bias in LSTM-LMs for learning syntax.
It is multi-task learning across the cross-entropy loss ($L_{lm}$) and an additional loss ($L_{add}$):
\begin{equation}\eqlabel{totalloss}
 L = L_{lm} + \beta L_{add},
\end{equation}
where $\beta$ is a relative weight for $L_{add}$.
Given outputs of LSTMs, a linear and binary softmax layers predict whether the next token is singular or plural.
$L_{add}$ is a loss for this classification, only defined for the contexts preceding a target token $x_{i}$:
\begin{equation*}
  L_{add} = \sum_{x_{1:i} \in \mathbf{h^*}} -\log p(\textrm{num}(x_{i}) | x_{1:i-1}),
\end{equation*}
where $x_{1:i} = x_1 \cdots x_{i}$ is a prefix sequence and $\mathbf{h^*}$ is a set of all prefixes ending with a target word (e.g., {\it An industrial park with several companies is}) in the training data.
$\textrm{num}(x) \in \{ \textrm{singular, plural} \}$ is a function returning the number of $x$.
In practice, for each mini-batch for $L_{lm}$, we calculate $L_{add}$ for the same set of sentences and add these two to obtain a total loss for updating parameters.

As we mentioned in Section~\secref{intro}, this loss does not exploit negative examples explicitly;
essentially a model is only informed of a key position (target word) that determines the grammaticality.
This is rather an indirect learning signal, and we expect that it does not outperform the other approaches.

\paragraph{Unlikelihood loss}
This is recently proposed \cite{welleck2019neural} for resolving the {\it repetition} issue, a known problem for neural text generators \cite{holtzman2019curious}.
Aiming at learning a model that can suppress repetition, they introduce an unlikelihood loss, which is an additional loss at a token level and explicitly penalizes choosing words previously appeared in the current context.

We customize their loss for negative tokens $x_i^*$ (e.g., {\it are} in (\ref{ex:incorrect_batch})).
Since this loss is added at token-level, instead of Eq.~\eqref{totalloss} the total loss is $L_{lm}$, which we modify as:
\begin{align}
 \sum_{\mathbf x \in D} \sum_{x_i \in \mathbf x} &-\log p(x_i|x_{1:i-1}) + \sum_{x_i^* \in \textrm{neg}_t(x_i)} g(x_i^*), \nonumber \\
 g(x_i^*) &= - \alpha \log (1 -p(x_i^* | x_{1:i-1})), \nonumber
\end{align}
where $\textrm{neg}_t(\cdot)$ returns negative tokens for a target $x_i$.\footnote{
Empty for non-target tokens.
It may return multiple tokens sometimes, e.g., themselves$\rightarrow$\{himself, herself\}.
}
$\alpha$ controls the weight.
$\mathbf x$ is a sentence in the training data $D$.
The unlikelihood loss strengthens the signal to penalize undesirable words in a context by explicitly reducing the likelihood of negative tokens $x_i^*$.
This is a more direct learning signal than the binary classification loss.

\paragraph{Sentence-level margin loss}
We propose a different loss, in which the likelihoods for correct and incorrect sentences are more tightly coupled.
As in the binary classification loss, the total loss is given by Eq.~\eqref{totalloss}.
We consider the following loss for $L_{add}$:
\begin{equation*}
 \sum_{\mathbf x \in D} \sum_{\mathbf x_j^* \in \textrm{neg}_s(\mathbf x)} \max(0, \delta - (\log p(\mathbf x) - \log p(\mathbf x_j^*))),
\end{equation*}
where $\delta$ is a margin value between the log-likelihood of original sentence $\mathbf x$ and negative sentences $\{ \mathbf x_j^* \}$.
$\textrm{neg}_s(\cdot)$ returns a set of negative sentences by modifying the original one.
Note that we change only one token for each $\mathbf x_j^*$, and thus may obtain multiple negative sentences from one $\mathbf x$ when it contains multiple target tokens (e.g., {\it she \underline{leaves} there but \underline{comes} back ...}).\footnote{
In principle, one can cumulate this loss within a single mini-batch for $L_{lm}$ as we do for the binary-classification loss.
However, obtaining $L_{add}$ needs to run an LM entirely on negative sentences as well, which demands a lot of GPU memories.
We avoid this by separating mini-batches for $L_{lm}$ and $L_{add}$.
We precompute all possible pairs of ($\mathbf x$, $\mathbf x_j^*$) and create a mini-batch by sampling from them.
We make the batch size for $L_{add}$ (the number of pairs) as the half of that for $L_{lm}$, to make the number of sentences contained in both kinds of batches equal.
Finally, in each epoch, we only sample at most the half mini-batches of those for $L_{lm}$ to reduce the total amount of training time.
}

Comparing to the unlikelihood loss, not only decreasing the likelihood of a negative example, this loss tries to guarantee a certain difference between the two likelihoods.
The learning signal of this loss seems stronger in this sense;
however, the token-level supervision is missing, which may provide a more direct signal to learn a clear contrast between correct and incorrect words.
This is an empirical problem we pursue in the experiments.

\paragraph{Token-level margin loss}
Our final loss is a combination of the previous two, by replacing $g(x_i)$ in the unlikelihood loss by a margin loss:
\begin{align}
 g(x_i^*) = \max(0, \delta-&(\log p(x_i|x_{1:i-1}) \nonumber \\
                 &- \log p(x_i^*|x_{1:i-1})). \nonumber
\end{align}
We will see that this loss is the most advantageous in the experiments (Section~\secref{exp}).

\subsection{Parameters}
Each method employs a few additional hyperparameters ($\beta$ for the binary classification loss, $\alpha$ for the unlikelihood loss, and $\delta$ for the margin losses).
We preliminary select $\beta$ and $\alpha$ from $\{1,10,100,1000\}$ that achieve the best average syntactic performance and find $\beta=1$ and $\alpha=1000$.
For the two margin losses, we fix $\beta=1.0$ and $\alpha=1.0$ and only see the effects of margin value $\delta$.

\begin{table*}
\centering
\scalebox{0.74}{
\begin{tabular}{lrrrrrrr}
\toprule
 & \multicolumn{2}{c}{LSTM-LM} & \multicolumn{2}{c}{Additional margin loss ($\delta=10$)} & \multicolumn{2}{c}{Additional loss {\small ($\alpha=1000, \beta=1$)}} & \multicolumn{1}{c}{Distilled} \\
 \cmidrule(lr){2-3} \cmidrule(lr){4-5} \cmidrule(lr){6-7} \cmidrule{8-8}
 & \multicolumn{1}{c}{M\&L18}&\multicolumn{1}{c}{Ours}&\multicolumn{1}{c}{Sentence-level}&\multicolumn{1}{c}{Token-level}&\multicolumn{1}{c}{Binary-pred.}&\multicolumn{1}{c}{Unlike.} & \multicolumn{1}{c}{K19} \\
\midrule
\textsc{Agreement}: &&&&&&\\
Simple & 94.0&98.1 ($\pm$1.3)&\textbf{100.0} ($\pm$0.0)&\textbf{100.0} ($\pm$0.0)&99.1 ($\pm$1.2)&99.7 ($\pm$0.6)&\textbf{100.0} ($\pm$0.0) \\
In a sent. complement & 99.0&96.1 ($\pm$2.0)&95.8 ($\pm$0.7)&\textbf{99.3} ($\pm$0.4)&96.9 ($\pm$2.4)&92.7 ($\pm$3.1)&98.0 ($\pm$2.0) \\
Short VP coordination & 90.0&93.6 ($\pm$3.0)&\textbf{100.0} ($\pm$0.0)&99.4 ($\pm$1.1)&93.8 ($\pm$3.3)&95.6 ($\pm$3.0)&99.0 ($\pm$2.0) \\
Long VP coordination & 61.0&82.2 ($\pm$3.4)&94.5 ($\pm$1.0)&\textbf{99.0} ($\pm$0.8)&83.9 ($\pm$3.2)&90.0 ($\pm$2.4)&80.0 ($\pm$2.0) \\
Across a PP & 57.0&92.6 ($\pm$1.4)&\textbf{98.8} ($\pm$0.4)&98.6 ($\pm$0.3)&92.7 ($\pm$1.3)&95.2 ($\pm$1.2)&91.0 ($\pm$3.0) \\
Across a SRC & 56.0&91.5 ($\pm$3.4)&99.6 ($\pm$0.4)&\textbf{99.8} ($\pm$0.2)&91.9 ($\pm$2.5)&97.1 ($\pm$0.7)&90.0 ($\pm$2.0) \\
Across an ORC & 50.0&84.5 ($\pm$3.1)&93.5 ($\pm$4.0)&\textbf{93.7} ($\pm$2.0)&86.3 ($\pm$3.2)&88.7 ($\pm$4.1)&84.0 ($\pm$3.0) \\
Across an ORC (no that) & 52.0&75.7 ($\pm$3.3)&86.7 ($\pm$4.2)&\textbf{89.4} ($\pm$2.7)&78.6 ($\pm$4.0)&86.4 ($\pm$3.5)&77.0 ($\pm$2.0) \\
In an ORC & 84.0&84.3 ($\pm$5.5)&99.8 ($\pm$0.2)&\textbf{99.9} ($\pm$0.1)&89.3 ($\pm$6.2)&92.4 ($\pm$3.5)&92.0 ($\pm$4.0) \\
In an ORC (no that) & 71.0&81.8 ($\pm$2.3)&97.0 ($\pm$1.0)&\textbf{98.6} ($\pm$0.9)&83.0 ($\pm$5.1)&88.9 ($\pm$2.4)&92.0 ($\pm$2.0) \\
\midrule
\textsc{Reflexive}: &&&&&\\
Simple & 83.0&94.1 ($\pm$1.9)&99.4 ($\pm$1.1)&\textbf{99.9} ($\pm$0.2)&91.8 ($\pm$2.9)&98.0 ($\pm$1.1)&91.0 ($\pm$4.0) \\
In a sent. complement & 86.0&80.8 ($\pm$1.7)&\textbf{99.2} ($\pm$0.6)&97.9 ($\pm$0.8)&79.0 ($\pm$3.1)&92.6 ($\pm$2.9)&82.0 ($\pm$3.0) \\
Across an ORC & 55.0&74.9 ($\pm$5.0)&72.8 ($\pm$2.4)&73.9 ($\pm$1.3)&72.3 ($\pm$3.0)&\textbf{78.9} ($\pm$8.6)&67.0 ($\pm$3.0) \\
\midrule
\textsc{NPI}: &&&&&\\
Simple & 40.0&\textbf{99.2} ($\pm$0.7)&98.7 ($\pm$1.6)&97.7 ($\pm$2.0)&98.0 ($\pm$3.1)&98.2 ($\pm$1.2)&94.0 ($\pm$4.0) \\
Across an ORC & 41.0&63.5 ($\pm$15.0)&56.8 ($\pm$6.0)&64.1 ($\pm$13.8)&64.5 ($\pm$14.0)&48.5 ($\pm$6.4)&\textbf{91.0} ($\pm$7.0) \\
\midrule
Perplexity & 78.6&\textbf{49.5} ($\pm$0.2)&56.4 ($\pm$0.5)&50.4 ($\pm$0.6)&49.6 ($\pm$0.3)&50.3 ($\pm$0.2)&56.7 ($\pm$0.2) \\
\bottomrule
\end{tabular}
}
 \caption{Comparison of syntactic dependency evaluation accuracies across different types of dependencies and perplexities.
 Numbers in parentheses are standard deviations.
 M\&L18 is the result of two-layer LSTM-LM in \citet{marvin-linzen:2018:EMNLP}.
 K19 is the result of distilled two-layer LSTM-LM from RNNGs \cite{kuncoro-etal-2019-scalable}.
 VP: verb phrase; PP: prepositional phrase; SRC: subject relative clause; and ORC: object-relative clause.
 Margin values are set to 10, which works better according to Figure~\figref{margin}.
 Perplexity values are calculated on the test set of the Wikipedia dataset.
 The values of M\&L18 and K19 are copied from \citet{kuncoro-etal-2019-scalable}.
 }\tablabel{main}
\end{table*}

\begin{figure*}[t]
 \centering
 \scalebox{0.97}{
 \includegraphics[width=\linewidth]{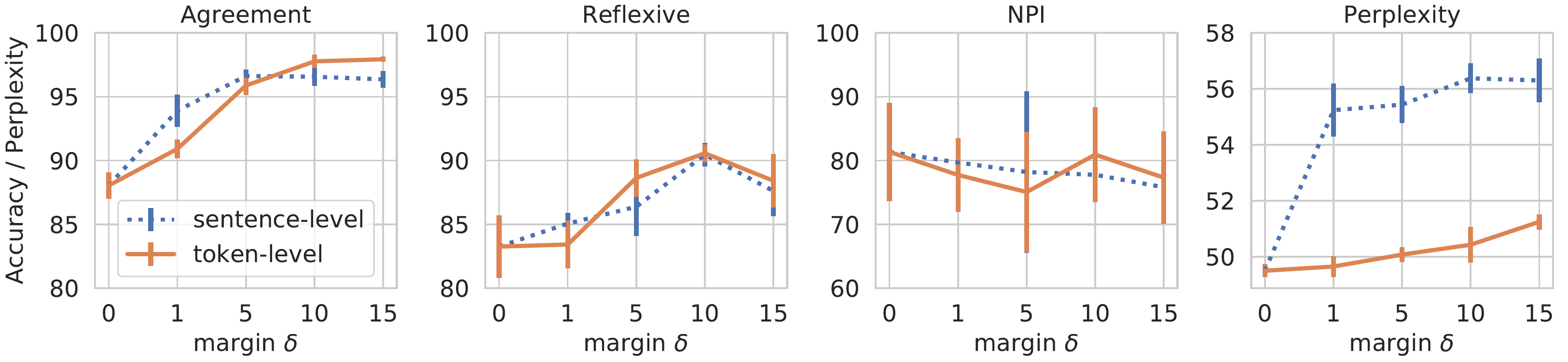}}
 \caption{Margin value vs.~macro average accuracy over the same type of constructions,
 or perplexity, with standard deviation for the sentence and token-level margin losses.
 $\delta=0$ is the baseline LSTM-LM without additional loss.
 }\figlabel{margin}
\end{figure*}

\subsection{Scope of Negative Examples}
\seclabel{scope}
Since our goal is to understand to what extent LMs can be sensitive to the target syntactic constructions by giving explicit supervision via negative examples, we only prepare negative examples on the constructions that are directly tested at evaluation.
Specifically, we mark the following words in the training data, and create negative examples:
\begin{description}
 \item[Present verb] To create negative examples on subject-verb agreement, we mark all present verbs and change their numbers.\footnote{
            We use Stanford tagger \cite{toutanova2003feature} to find the present verbs.
            We change the number of verbs tagged by VBZ or VBP using \texttt{inflect.py} ({\color{darkblue}https://pypi.org/project/inflect/}).}
 \item[Reflexive pronoun] We also create negative examples on reflexive anaphora, by flipping between \{\textit{themselves}\}$\leftrightarrow$\{\textit{himself, herself}\}.
\end{description}
These two are both related to the syntactic number of a target word.
For binary classification we regard both as a target word, apart from the original work that only deals with subject-verb agreement \cite{enguehard-etal-2017-exploring}.
We use a single common linear layer for both constructions.

In this work, we do not create negative examples for NPIs.
This is mainly for technical reasons.
Among four losses, only the sentence-level margin loss can correctly handle negative examples for NPIs, essentially because other losses are token-level.
For NPIs, left contexts do not have information to decide the grammaticality of the target token (a quantifier; no, most, etc.) (Section~\secref{task}).
Instead, in this work, we use NPI test cases as a proxy to see possible negative (or positive) impacts as compensation for specially targeting some constructions.
We will see that in particular for our margin losses, such negative effects are very small.

\section{Experiments on Additional Losses}
\seclabel{exp}

We first see the overall performance of baseline LSTM-LMs as well as the effects of additional losses.
Throughout the experiments, for each setting, we train five models from different random seeds and report the average score and standard deviation.
The code is available at {\color{darkblue}https://github.com/aistairc/lm\_syntax\_negative}.

\paragraph{Naive LSTM-LM performs well}
The main accuracy comparison across target constructions for different settings is presented in Table~\tabref{main}.
We first notice that our baseline LSTM-LM (Section~\secref{lm}) performs much better than \citet{marvin-linzen:2018:EMNLP}'s LM.
A similar observation is recently made by \citet{kuncoro-etal-2019-scalable}.\footnote{We omit the comparison but the scores are overall similar.}
This suggests that the original work underestimates the true syntactic ability induced by LSTM-LMs.
The table also shows the results by their distilled LSTM-LM from RNNGs (Section~\secref{intro}).

\begin{figure*}[t]
 \centering
 \scalebox{0.97}{
 \includegraphics[width=\linewidth]{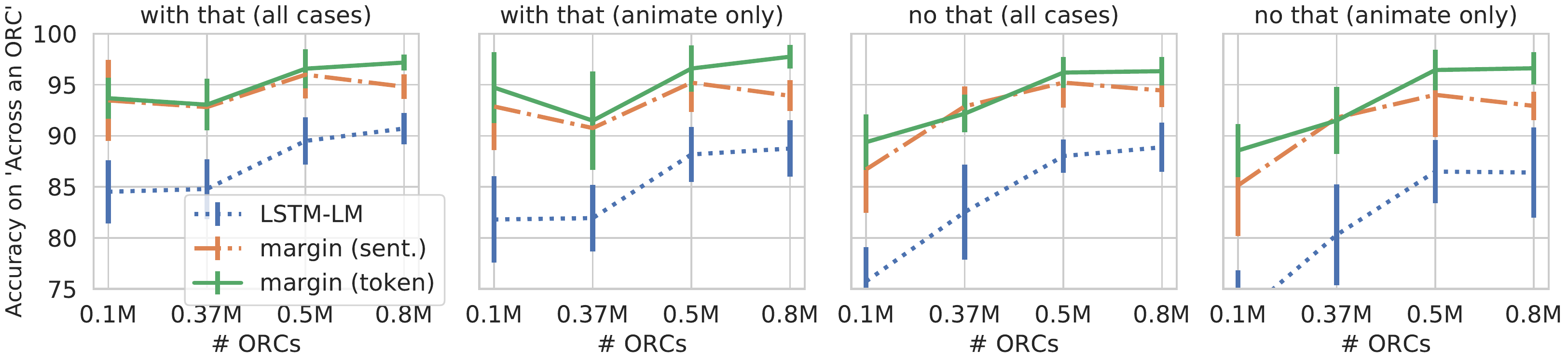}}
 \caption{Accuracies on ``Across an ORC'' (with and without complementizer ``that'') by models trained on augmented data with additional sentences containing an object RC.
 Margin is set to 10.
 X-axis denotes the total number of object RCs in the training data.
 0.37M roughly equals the number of subject RCs in the original data.
 ``animate only'' is a subset of examples (see body).
 Error bars are standard deviations across 5 different runs.
 }\figlabel{orc}
\end{figure*}

\paragraph{Higher margin value is effective}
For the two types of margin loss, which margin value should we use?
Figure~\figref{margin} reports average accuracies within the same types of constructions.
For both token and sentence-levels, the task performance increases along $\delta$, but a too large value (15) causes a negative effect, in particular on reflexive anaphora.
Increases (degradations) of perplexity are observed in both methods but this effect is much smaller for the token-level loss.
In the following experiments, we fix the margin value to 10 for both, which achieves the best syntactic performance.

\paragraph{Which additional loss works better?}
We see a clear tendency that our token-level margin loss achieves overall better performance.
Unlikelihood loss does not work unless we choose a huge weight parameter ($\alpha=1000$), but it does not outperform ours, with a similar value of perplexity.
The improvements by binary-classification loss are smaller, indicating that the signals are weaker than other methods with explicit negative examples.
Sentence-level margin loss is conceptually advantageous in that it can deal with any type of sentence-level grammaticality including NPIs.
We see that it is overall competitive with token-level margin loss but suffers from a larger increase of perplexity (4.9 points), which is observed even with smaller margin values (Figure~\figref{margin}).
Understanding the cause of this degradation as well as alleviating it is an important future direction.

\section{Limitations of LSTM-LMs}
\seclabel{orc}
In Table~\tabref{main}, the accuracies on dependencies across an object RC are relatively low.
The central question in this experiment is whether this low performance is due to the limitation of current architectures, or other factors such as frequency.
We base our discussion on the contrast between object (\ref{ex:author_orc}) and subject (\ref{ex:author_src}) RCs:
\vspace{2pt}\begin{exe}
 \ex\label{ex:author_orc}The authors (that) the chef likes \underline{laugh}.
\end{exe}
\begin{exe}
 \ex\label{ex:author_src}The authors that like the chef \underline{laugh}.
\end{exe}
\vspace{2pt}
Importantly, the accuracies for a subject RC are more stable, reaching 99.8\% with the token-level margin loss, although the content words used in the examples are common.\footnote{\label{foot:animate}
Precisely, they are not the same.
Examples of object RCs are divided into two categories by the animacy of the main subject ({\it animate} or not), while subject RCs only contain animate cases.
If we select only animate examples from object RCs the vocabularies for both RCs are the same,
remaining only differences in word order and inflection, as in (\ref{ex:author_orc}, \ref{ex:author_src}).
}

It is known that object RCs are less frequent than subject RCs \cite{hale-2001-probabilistic,Levy2008-LEVESC}, and it could be the case that the use of negative examples still does not fully alleviate this factor.
Here, to understand the true limitation of the current LSTM architecture, we try to eliminate such other factors as much as possible under a controlled experiment.

\paragraph{Setup}
We first inspect the frequencies of object and subject RCs in the training data, by parsing them with the state-of-the-art Berkeley neural parser \cite{kitaev-klein:2018:Long}.
In total, while subject RCs occur 373,186 times, object RCs only occur 106,558 times.
We create three additional training datasets by adding sentences involving object RCs to the original Wikipedia corpus (Section~\secref{lm}).
To this end, we randomly pick up 30 million sentences from Wikipedia (not overlapped to any sentences in the original corpus), parse by the same parser, and filter sentences containing an object RC, amounting to 680,000 sentences.
We create augmented training sets by adding a subset, or all of these sentences to the original training sentences.
Among the test cases about object RCs we only report accuracies on subject-verb agreement, on which the portion for subject RCs also exists.
This allows us to compare the difficulties of two types of RCs for the present models.
We also evaluate on ``animate only'' subset, which has a correspondence to the test cases for subject RCs with only differences in word order and inflection (like (\ref{ex:author_orc}) and (\ref{ex:author_src}); see footnote \ref{foot:animate}).
Of particular interest to us is the accuracy on these animate cases.
We expect that the main reason for lower performance for object RCs is due to frequency, and with our augmentation the accuracy will reach the same level as that for subject RCs.

\begin{figure}[t]
 \centering
 \scalebox{0.9}{
 \includegraphics[width=\linewidth]{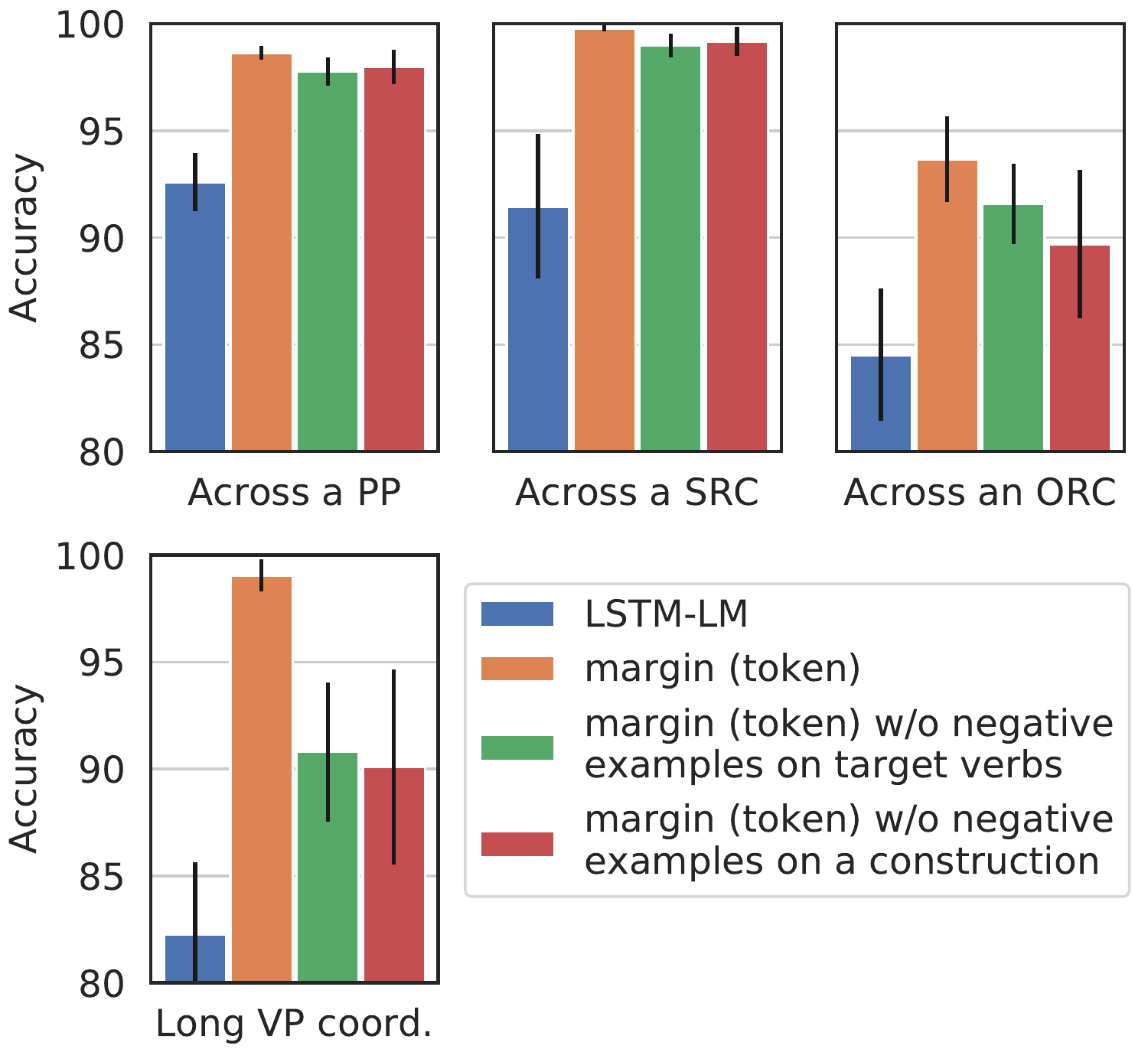}}
 \caption{An ablation study to see the performance of models trained with reduced explicit negative examples (token-level and construction-level).
 One color represents the same models across plots, except the last bar (construction-level), which is different for each plot.}\figlabel{ablation}
\end{figure}

\paragraph{Results}
However, for both all and animate cases, accuracies are below those for subject RCs (Figure~\figref{orc}).
Although we see improvements from the original score (93.7), the highest average accuracy by the token-level margin loss on the ``animate'' subset is 97.1 (``with that''), not beyond 99\%.
This result indicates some architectural limitations of LSTM-LMs in handling object RCs robustly at a near perfect level.
Answering why the accuracy does not reach (almost) 100\%, perhaps with other empirical properties or inductive biases \cite{khandelwal-etal-2018-sharp,ravfogel-etal-2019-studying} is future work.

\section{Do models generalize explicit supervision, or just memorize it?}

One distinguishing property of our margin loss, in particular token-level loss, is that it is highly lexical, making a contrast explicitly between correct and incorrect words.
This direct signal may make models acquire very specialized knowledge about each target word, not very generalizable one across similar words and occurring contexts.
In this section, to get insights into the transferability of syntactic knowledge induced by our margin losses, we provide an ablation study by removing certain negative examples during training.

\paragraph{Setup}
We perform two kinds of ablation.
For token-level ablation (\textsc{-Token}), we avoid creating negative examples for all verbs that appear as a target verb\footnote{{\it swim, smile, laugh, enjoy, hate, bring, interest, like, write, admire, love, know}, and {\it is}.} in the test set.
Another is construction-level (\textsc{-Pattern}), by removing all negative examples occurring in a particular syntactic pattern.
We ablate a single construction at a time for \textsc{-Pattern}, from four non-local subject-verb dependencies (across a prepositional phrase (PP), subject RC, object RC, and long verb phrase (VP)).\footnote{We identify all these cases from the parsed training data, which we prepared for the analysis in Section~\secref{orc}.}
We hypothesize that models are less affected by token-level ablation, as knowledge transfer across words appearing in similar contexts is promoted by language modeling objective.
We expect that construction-level supervision would be necessary to induce robust syntactic knowledge, as perhaps different phrases, e.g., a PP and a VP, are processed differently.

\begin{table}
\centering
\scalebox{0.78}{
\begin{tabular}{lccc}
\toprule
& \multicolumn{3}{c}{Second verb (V1 and \textbf{V2})} \\
Models & All verbs & like & other verbs \\
\midrule
LSTM-LM & 82.2 ($\pm$3.4)&13.0 ($\pm$12.2)&89.9 ($\pm$3.6) \\
Margin (token) & 99.0 ($\pm$0.8)&94.0 ($\pm$6.5)&99.6 ($\pm$0.5) \\
~~ \textsc{-Token} & 90.8 ($\pm$3.3)&51.0 ($\pm$29.9)&95.2 ($\pm$2.6) \\
~~ \textsc{-Pattern} & 90.1 ($\pm$4.6)&50.0 ($\pm$30.6)&94.6 ($\pm$2.2) \\
\bottomrule
\end{tabular}
}
 \caption{Accuracies on long VP coordinations by the models with/without ablations.
 ``All verbs'' scores are overall accuracies.
 ``like'' scores are accuracies on examples on which the second verb (target verb) is {\it like}.
 }\tablabel{vp_coord_first}
\end{table}
\begin{table}[t]
\centering
\scalebox{0.78}{
\begin{tabular}{lcc}
\toprule
& \multicolumn{2}{c}{First verb (\textbf{V1} and V2)} \\
Models & likes & other verbs \\
\midrule
LSTM-LM & 61.5 ($\pm$20.0)& 93.5 ($\pm$3.4) \\
Margin (token) & 97.0 ($\pm$4.5)&99.9 ($\pm$0.1) \\
~~ \textsc{-Token} &63.5 ($\pm$18.5)& 99.2 ($\pm$1.1) \\
~~ \textsc{-Pattern} &67.0 ($\pm$21.2)& 98.0 ($\pm$1.4) \\
\bottomrule
\end{tabular}
}
 \caption{Further analysis of accuracies on the ``other verbs'' cases of Table~\tabref{vp_coord_first}.
 Among these cases, the second column (``likes'') shows accuracies on examples where the first verb (not target) is {\it likes}.
 }\tablabel{second_vp}
\end{table}

\paragraph{Results}
Figure~\figref{ablation} is the main results.
Across models, we restrict the evaluation on four non-local dependency constructions, which we select as ablation candidates as well.
For a model with \textsc{-Pattern}, we evaluate only on examples of construction ablated in training (see caption).
To our surprise, both \textsc{-Token} and \textsc{-Pattern} have similar effects, except ``Across an ORC'', on which the degradation by \textsc{-Pattern} is larger.
This may be related to the inherent difficulty of object RCs for LSTM-LMs that we verified in Section~\secref{orc}.
For such particularly challenging constructions, models may need explicit supervision signals.
We observe lesser score degradation by ablating prepositional phrases and subject RCs.
This suggests that, for example, the syntactic knowledge strengthened for prepositional phrases with negative examples could be exploited to learn the syntactic patterns about subject RCs, even when direct learning signals on subject RCs are missing.

We see approximately 10.0 points score degradation on long VP coordination by both ablations.
Does this mean that long VPs are particularly hard in terms of transferability?
We find that the main reasons for this drop, relative to other cases, are rather technical, essentially due to the target verbs used in the test cases.
See Table~\tabref{vp_coord_first,second_vp}, which show that failed cases for the ablated models are often characterized by the existence of either {\it like} or {\it likes}.
Excluding these cases (``other verbs'' in Table~\tabref{second_vp}), the accuracies reach 99.2 and 98.0 by \textsc{-Token} and \textsc{-Pattern}, respectively.
These verbs do not appear as a target verb in the test cases of other tested constructions.
This result suggests that the transferability of syntactic knowledge to a particular word may depend on some characteristics of that word.
We conjecture that the reason for weak transferability to {\it likes} and {\it like} is that they are polysemous;
e.g., in the corpus, {\it like} is much more often used as a preposition and being used as a present tense verb is rare.
This type of issue due to frequency may be one reason for lessening the transferability.
In other words, {\it like} can be seen as a challenging verb to learn its usage only from the corpus, and our margin loss helps for such cases.

\section{Discussion and Conclusion}
Our results with explicit negative examples are overall positive.
We have demonstrated that models exposed to these examples at training time in an appropriate way will be capable of handling the targeted constructions at near perfect level except a few cases.
We found that our new token-level margin loss is superior to the other approaches and the remaining challenging cases are dependencies across an object relative clause.

Object relative clauses are known to be harder for a human as well, and our results may indicate some similarities in the sentence processing behaviors by a human and RNN, though other studies also find some dissimilarities between them \cite{linzen2018distinct,wilcox-etal:2019-what-syntactic-structures}.
The difficulty of object relative clauses for RNN-LMs has also been observed in the prior work \cite{marvin-linzen:2018:EMNLP,van-schijndel-EtAl:2019:EMNLP1}.
A new insight provided by our study is that this difficulty holds even after alleviating the frequency effects by augmenting the target structures along with direct supervision signals.
This indicates that RNNs might inherently suffer from some memory limitation like a human subject, for which the difficulty of particular constructions, including center-embedded object relative clauses, are known to be incurred due to memory limitation \cite{gibson1998linguistic,demberg-keller-2008} rather than purely frequencies of the phenomena.
In terms of language acquisition, the supervision provided in our approach can be seen as direct negative evidence \cite{MARCUS199353}.
Since human learners are known to acquire syntax without such direct feedback we do not claim that our proposed learning method itself is cognitively plausible.

One limitation of our approach is that the scope of negative examples has to be predetermined and fixed.
Alleviating this restriction is an important future direction.
Though it is challenging, we believe that our final analysis for transferability, which indicates that the negative examples do not have to be complete and can be noisy, suggests a possibility of a mechanism to induce negative examples themselves during training, perhaps relying on other linguistic cues or external knowledge.

\section*{Acknowledgements}
We would like to thank Naho Orita and the members of Computational Psycholinguistics Tokyo for their valuable suggestions and comments.
This paper is based on results obtained from projects commissioned by the New Energy and Industrial Technology Development Organization (NEDO).

\bibliography{main}
\bibliographystyle{acl_natbib}

\end{document}